\documentclass{article} 
\usepackage{GEM_workshop_2025, times}

\usepackage{amsmath,amsfonts,bm}

\def\eqref#1{equation~\ref{#1}}

\def\1{\bm{1}}

\DeclareMathAlphabet{\mathsfit}{\encodingdefault}{\sfdefault}{m}{sl}
\SetMathAlphabet{\mathsfit}{bold}{\encodingdefault}{\sfdefault}{bx}{n}

\usepackage{hyperref}
\usepackage{url}
\usepackage{graphicx}

\usepackage[ruled,vlined]{algorithm2e}
\usepackage{graphicx}
\usepackage{caption}
\usepackage{comment}
\usepackage{float}

\title{Active Learning on Synthons for Molecular Design}

\makeatletter
\renewcommand*{\@fnsymbol}[1]{\ensuremath{\ifcase#1\or \dagger\or  *\or  \ddagger\or
   \mathsection\or \mathparagraph\or \|\or **\or \dagger\dagger
   \or \ddagger\ddagger \else\@ctrerr\fi}}
\makeatother

\newcommand{\sep}{\,\,\,}

\author{Tom George Grigg$^1$\thanks{Equal contribution} \sep Mason Burlage$^1$\footnotemark[1] \sep Oliver Brook Scott$^1$\sep \textbf{Adam Taouil}$^1$\sep \textbf{Dominique Sydow}$^1$\sep \\ \textbf{Liam Wilbraham}$^1$ \\
$^1$Recursion \\ \texttt{liam.wilbraham@recursion.com} \\
}

\iclrfinalcopy 
\begin{document}

\maketitle
\vspace{-10px}

\begin{abstract}
Exhaustive virtual screening is highly informative but often intractable against the expensive objective functions involved in modern drug discovery. 
This problem is exacerbated in combinatorial contexts such as multi-vector expansion, where molecular spaces can quickly become ultra-large. Here, we introduce Scalable Active Learning via Synthon Acquisition (SALSA): a simple algorithm applicable to multi-vector expansion which extends pool-based active learning to non-enumerable spaces by factoring modeling and acquisition over synthon or fragment choices. Through experiments on ligand- and structure-based objectives, we highlight SALSA's sample efficiency, and its ability to scale to spaces of trillions of compounds. Further, we demonstrate application toward multi-parameter objective design tasks on three protein targets – finding SALSA-generated molecules have comparable chemical property profiles to known bioactives, and
exhibit greater diversity and higher scores over an industry-leading generative approach.
\end{abstract}

\section{Introduction}

Given the strong association between a molecule's core scaffold and its chemical properties, a common workflow is to iteratively design, make, and test changes at targeted R-groups in order to advance therapeutics through the discovery pipeline \citep{design_automation}. Exhaustive virtual screening of R-group changes aids designers and medicinal chemists in the search for promising, synthesizable molecular structures, but quickly becomes intractable against computationally expensive scores as the number of possible attachments increases. Prior work in MolPAL \citep{Graff} extends the scope of screening to spaces on the order of 100M molecules via pool-based deep Bayesian optimization. However, in the context of multi-vector expansion, where multiple R-groups are explored simultaneously, spaces can easily surpass 100B+ possible combinations in early stage discovery. At this scale,  designers often turn to specialized cheminformatics tools which can be configured to screen constrained synthesizable spaces for substructure \citep{fast-substructure},  similarity \citep{fast-similarity, SASS}, and docking-based \citep{Sadybekov} design objectives. 

For bespoke or multi-parameter objectives (MPOs), designers may employ generative (or \textit{inverse}) design. Modern generative approaches typically optimize pre-trained prior distributions on graphs or SMILES towards a molecular score e.g. via RL \citep{Reinvent}, guided diffusion \citep{guided_diffusion}, etc. Historically, these methods faced issues with synthetic accessibility \citep{Gao, Gen_failures}, but recent works mitigate with explicit synthesis constraints \citep{de_novo_syn, gen-syn-mols, Libinvent}, or analogizing \citep{analogue-search, syn-gen-analogue}. However, their usage remains limited in practice due to the unwieldy tension between synthesizability and drug-likeness versus novelty when sampling from a generative molecular model.

Here, we extend the domain of pool-based active learning (AL) to a multi-vector expansion context by introducing \textbf{S}calable \textbf{A}ctive \textbf{L}earning via \textbf{S}ynthon \textbf{A}cquisition (SALSA). By factoring learning over independent synthon or fragment choices, SALSA facilitates screening in explicitly configurable molecular spaces on the order of trillions of compounds. We demonstrate that SALSA is sample efficient with respect to baselines, and validate its application to multi-vector design tasks on three protein targets. We find that SALSA identifies molecules with comparable chemical property distributions to known bioactive compounds while optimizing pharmacophore and structure-based MPOs -- improving on established generative approaches, and offering a pragmatic alternative.

\begin{figure}
    \centering
    \includegraphics{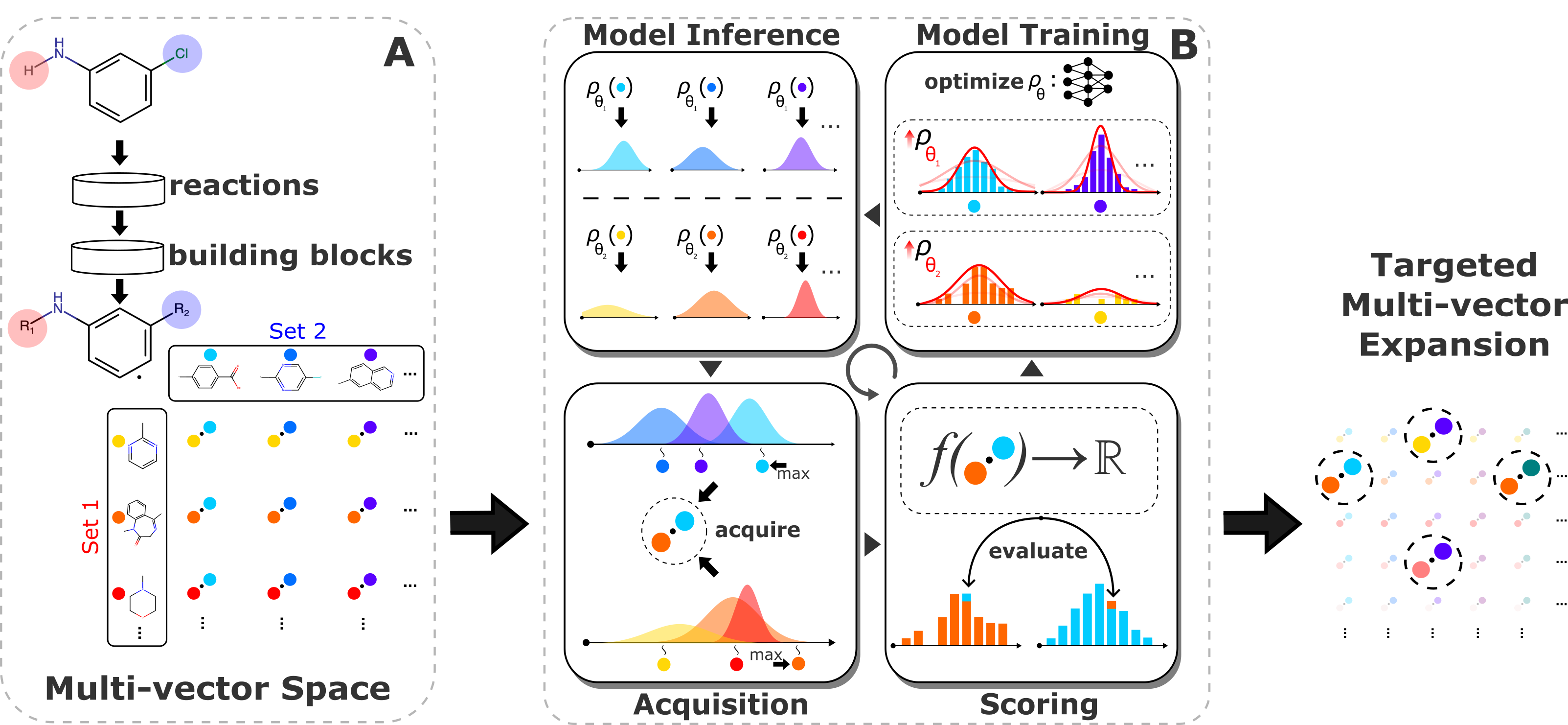}
    \captionsetup{labelfont=bf}
    \caption{\textbf{A} Construction of a 2-vector synthon space. \textbf{B} AL loop against scoring function $f$.
    }
    \label{fig:overview}
\end{figure}

\section{Methods}

\paragraph{Search space}
SALSA consumes as input a target molecular space formed by pre-defined choices of synthons or fragments, as well as a molecular objective function $f$. Fig. \ref{fig:overview}A exemplifies construction of a target space for a simple 2-vector expansion scheme on a core with two R-groups. Given a set of SMIRKS-encoded reactions and building blocks, applicable chemistry is represented by a synthon set $\mathcal{S}_i$ for each vector, determined by efficient pattern matching. In our experiments, we use building blocks from \cite{Mcule}, and a set of custom SMIRKS (see \ref{appendix:synthon_spaces}).

\vspace{-1mm}

\paragraph{Synthon acquisition}
Fig. \ref{fig:overview}B illustrates SALSA for 2-vector expansion. $K$ molecules are initially sampled randomly and scored with $f$. Scores are recorded for each molecule's constituent synthons, and a surrogate model is trained at each vector. Score distributions are predicted for all synthons, and $K$ new molecules are sampled via an acquisition strategy $\alpha$. These molecules are scored to form additional synthon datapoints with which to retrain. This process loops for $N$ rounds, or until convergence – acquiring up to $N \times K$ molecules. Conceptually, SALSA navigates two (or $n$, in general) non-stationary multi-armed bandit problems simultaneously, one for the pool of synthons at each vector. Factoring the decision problem in this way nullifies inference-time combinatorial complexity from $O(\prod_{i} |\mathcal{S}_i|)$ to $O(\sum(|\mathcal{S}_i)|)$, which is the primary limitation for full-molecular AL.

In our experiments, we adapt the (approximate) Thompson sampling (TS) strategy outlined in MolPAL \citep{Graff} to the multi-vector case. Here, synthon acquisition scores are sampled from a predicted Gaussian, i.e. $\alpha(s) \sim \mathcal{N}(\mu_\theta(s), \sigma_\theta(s)) \,\, \forall s \in \mathcal{S}_i$ for learned parameters $\theta$. Top-scoring synthons are combined and the resulting molecule is scored if unseen, otherwise synthons are resampled. The loop terminates early if more than a given threshold of samples are rejected in a round – as this suggests convergence in acquisition probability. We include pseudo-code describing SALSA in \ref{appendix:algorithm}. We also adapted and investigated alternative acquisition strategies (see \ref{appendix:alternative_acquisition_strategies}), as well as a variant that uses one model for all synthon sets (\ref{appendix:one_model_vs_two}), rather than a model per vector.

\paragraph{Surrogate models}
We adopt \texttt{chemprop}'s  \citep{chemprop} implementation of a directed message-passing neural network (MPNN) as our choice of surrogate model. The MPNN dynamically encodes a feature vector by aggregating rounds of message passing across the bonds of a molecule's 2D graph. A feed-forward head with two output nodes then operates on this graph-based representation to predict a mean $\mu_\theta(s)$ and variance $\sigma_\theta(s)$ for a synthon $s$. We train these models end-to-end to predict synthon score distributions via a mean-variance estimation (MVE) loss  $\mathcal{L}(y, s, \theta) = \frac{\log 2\pi}{2} + \log{\sigma_\theta(s)} - \frac{1}{2}\left(\frac{y - \mu_\theta(s)}{\sigma_\theta(s)}\right)^2$ – i.e. maximum-likelihood estimation of Gaussian density for observed synthon-score pairs $(s, y) \in \mathcal{S}_i \times \mathbb{R}$ (up to regularization induced priors). We found this model to perform better than fixed-feature alternatives (\ref{appendix:alternative_acquisition_strategies}), consistent with findings in MolPAL. Details on architecture and hyperparameter choices are included in \ref{appendix:MPNN_hyperparameters}.

\section{Experiments}\label{section:results}

\begin{figure}
    \centering
\includegraphics{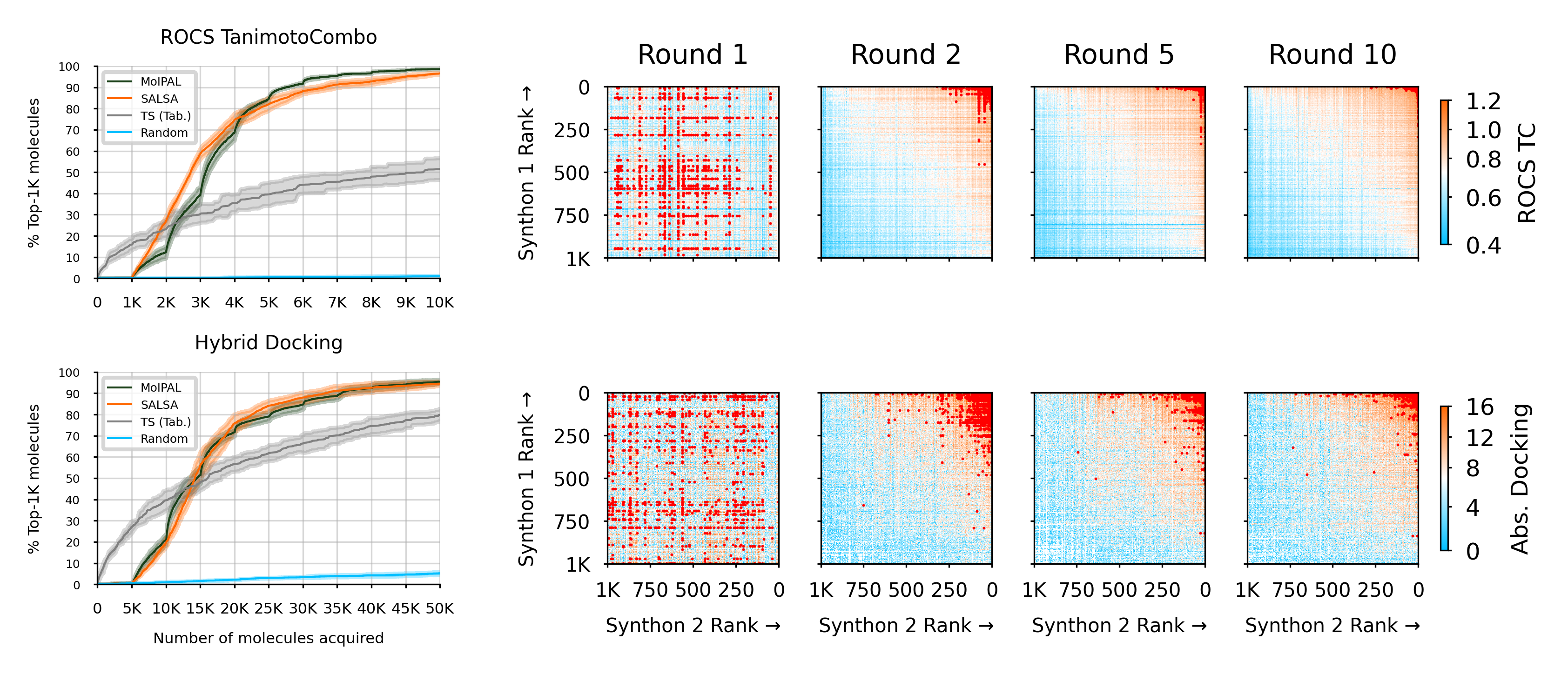}
    \captionsetup{labelfont=bf}
    \caption{\label{fig:benchmark} Recall of top-1K compounds in the 1M target space for ROCS-TC (top) and docking (bottom) as a function of molecules acquired, smoothed over 5 trials. The heatmaps show the enumerated target space decomposed across synthon axes and coloured by score. For a given SALSA round, synthons are ordered by Monte Carlo-estimated acquisition probability, i.e. the likelihood of sampling increases moving up and right. Top-1K ground truth molecules are highlighted in red.}
\end{figure}

\paragraph{Sample efficiency}
We begin by demonstrating SALSA's performance in an enumerated 1M molecule space, using CDK2 as a model system (PDB: 6GUH \citep{CDK2}). To set-up our expansion, we isolated the co-crystallized ligand's core and functionalized reaction handles at two vector positions. Synthons were generated at each vector (see \ref{appendix:synthon_spaces}), resulting in 910K and 2.4M synthons respectively. 1K synthons were subsampled from each set to create a 1K$\times$1K=1M-size space as desired. We defined a docking score using the protein structure, and a 3D shape/color similarity score using the co-crystallized ligand as a reference via OpenEye Hybrid Docking and ROCS TanimotoCombo Score (ROCS-TC), respectively (see \ref{appendix:scores}). The space was exhaustively enumerated and scored with both objectives to obtain ground truth, enabling comparison of SALSA to: TS (Tab.) \citep{Klarich} which is conceptually similar but updates tabular Gaussian models with exact fixed-variance TS, and a MolPAL-like \citep{Graff} full-molecular pool-based screen.

Fig. \ref{fig:benchmark} depicts 10 rounds of SALSA for each task, with an objective scoring budget of 1K and 5K molecules per round for ROCS-TC and docking, respectively. SALSA is able to ultimately identify 96.5\% and 94.5\% of the top-1K molecules, greatly outperforming random screening. Importantly, MolPAL retrieves 98.5\% and 95.4\% of the top-1K given the same model configuration and budget – revealing that factored synthon acquisition degrades performance minimally compared to learning over full molecules for these tasks. Interestingly, SALSA learns faster than MolPAL in early rounds for ROCS-TC, perhaps reflective of the approximate additivity of shape-based scoring across fragments \citep{SASS}. TS (Tab.) is significantly less sample efficient due to its lack of generalization across synthons. The heatmaps in Fig. \ref{fig:benchmark} illustrate SALSA's progressive ranking of the target space based on the probability of acquiring a given molecule's constituent synthons. SALSA quickly learns to separate high-scoring and low-scoring molecules, and refines its sampling distribution over multiple rounds to prioritise top molecules, visualised as the top-1K molecules (in red) moving steadily towards the top right corner where acquisition probability is highest.

\begin{figure}
    \centering
    \includegraphics{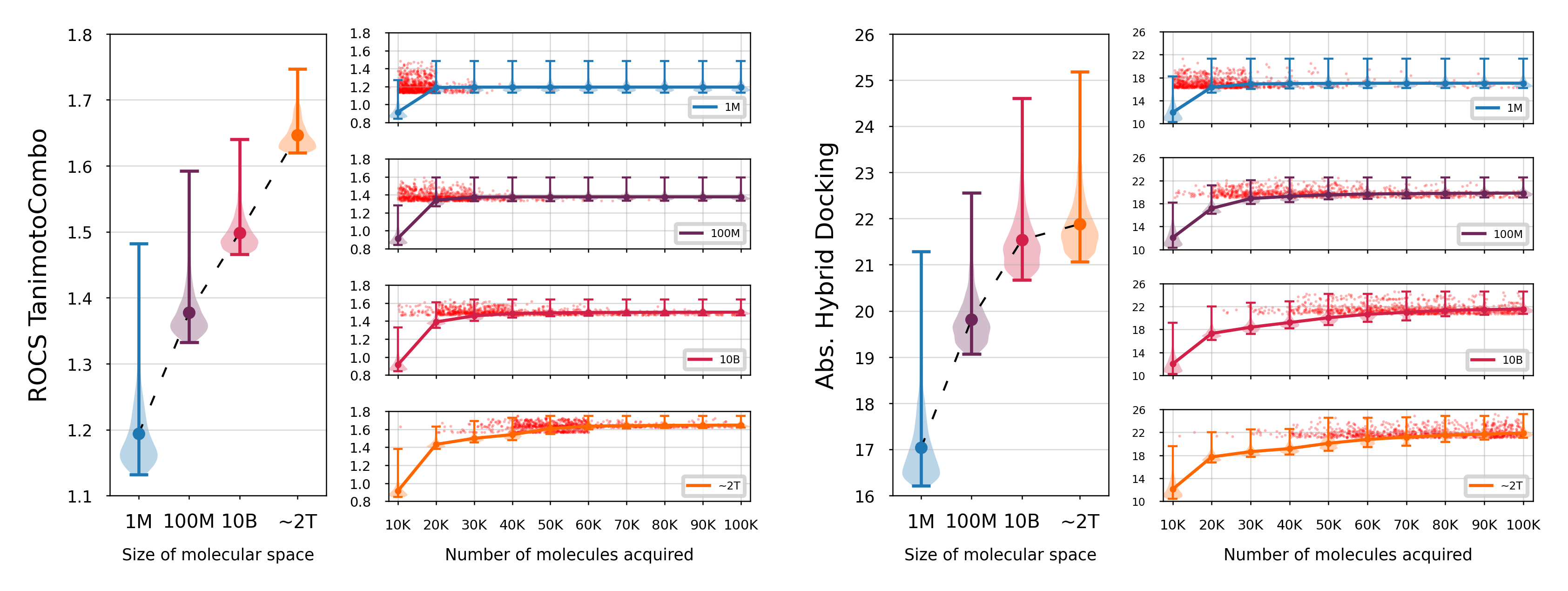}
    \captionsetup{labelfont=bf}
    \caption{\label{fig:scaling} The large violin plots show the min, max, mean, and estimated score density for the top-1K molecules identified by SALSA as space size increases for shape- (left) and structure-based (right) objectives, smoothed over 3 trials. Subplots on the right show the evolution of the top-1K distribution over AL rounds – the final top-1K molecules are marked in red at their sample index.}
\end{figure}

 \paragraph{Scaling beyond enumerable spaces} \label{experiments:sample_efficiency}
Next, we assess SALSA's ability to scale to multi-vector spaces beyond the domain of exhaustive screening. We use the same task setup as above, this time subsampling 1K, 10K, and 100K synthons for each vector to construct increasingly large spaces of size 1M, 100M, 10B, and a final $\sim$2T=910K$\times$2.4M space, where all synthons are made available. We fix our budget in each space for calls to both the Hybrid Docking and ROCS-TC objective functions to a reasonable 10K molecules per round for 10 rounds of active learning. For completeness, we report runtime for each space in \ref{appendix:runtime}.
  
Fig. \ref{fig:scaling} demonstrates that SALSA consistently finds better scoring molecules with increasing space size, where we observe an approximately log-linear improvement in ROCS-TC score. The rate of improvement appears to decrease for docking between 10B and $\sim$2T molecules. It is informative to look at the acquisition of top-1K molecules (in red) identified during a given run. For ROCS-TC, even the $\sim$2T space requires only 60K-70K sampled molecules before diminishing returns. In contrast, many new top molecules appear in the final rounds for docking in both the 10B and $\sim$2T spaces. This may indicate either model saturation, or that a 100K learning budget is insufficient to fully explore – consistent with Fig. \ref{fig:benchmark}B where learning converges more slowly for docking. However, the upper tail of the top-1K distribution appears to have converged, suggesting the possibility that there are few significantly higher scoring molecules identifiable by SALSA. \citet{Lyu_virtual_screening} demonstrate a log-linear improvement for top docking scores in increasingly large virtual screens, but in the multi-vector case the core is fixed, potentially constraining the highest achievable score. 

\begin{figure}
    \centering
\includegraphics[scale=0.16]{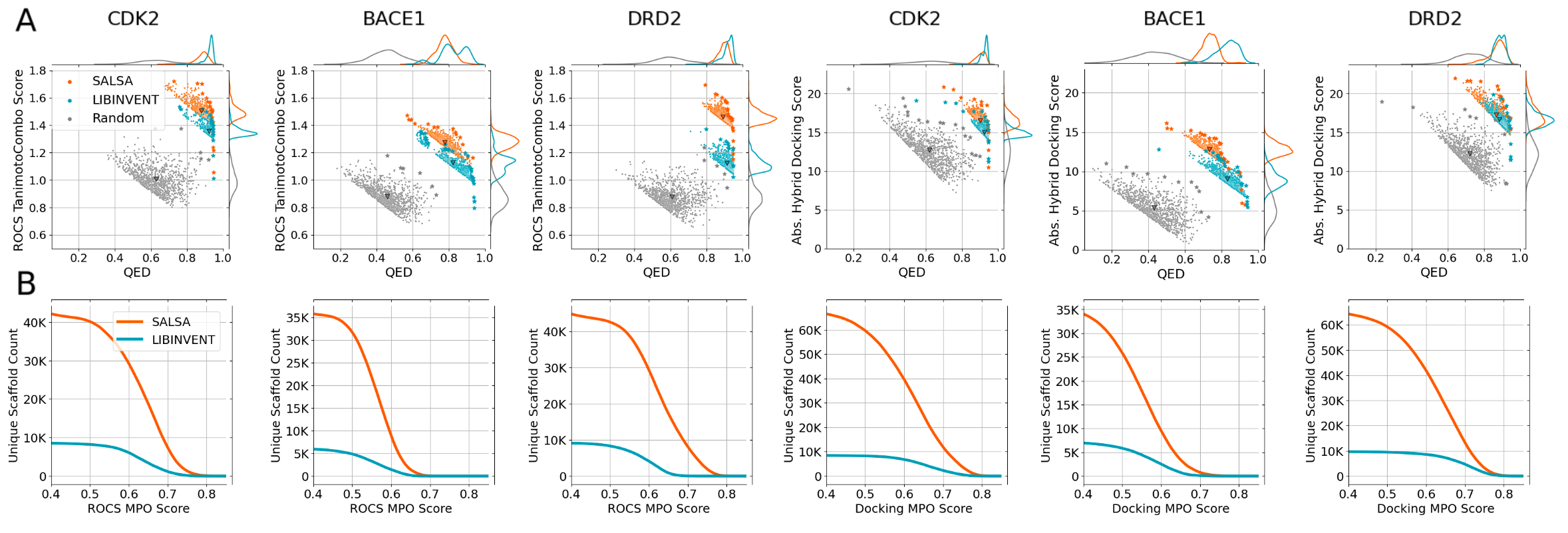}
    \captionsetup{labelfont=bf}
    \caption{\label{fig:three_proteins}
\textbf{A.} Top-1K molecules identified across three targets for SALSA, random acquisition, and LibINVENT using QED \citep{Bickerton} plus ROCS-TC and Hybrid Docking objectives. Mean scores and top-20 pareto optimal molecules are denoted by triangles and stars, respectively. \textbf{B.} Number of unique Bemis-Murcko scaffolds above a given score value for molecules identified by SALSA and LibINVENT. SALSA compounds show substantially greater diversity.}
\end{figure}

\paragraph{Multi-parameter objectives}
Finally, we explore SALSA's ability to optimize simple MPOs derived for targets from three protein classes: CDK2, a kinase; BACE1 (PDB: 2IRZ \citep{BACE1}), a protease; and DRD2 (PDB: 6LUQ \citep{D2}), a GPCR. Molecular cores were again extracted from co-crystallized ligands and used to generate synthons (see \ref{appendix:synthon_spaces}). We additionally impose light substructure filtering with standard structural alerts to represent a more realistic target space. After filtering, 547K$\times$1.3M, 1.0M$\times$1.1M, and 1.2M$\times$246K synthons remained for CDK2, BACE1, and DRD2, respectively. Two simple linear MPOs were assessed for each system: Hybrid Docking + QED, and ROCS-TC + QED, with each component scaled to $(0, 1)$, and a 1:1 and 2:1 weighting applied, respectively (see \ref{appendix:scores}). 10K objective function calls were again budgeted for each of 10 rounds. We plot the MPO components of the top-1K molecules acquired via random acquisition and SALSA in Fig. \ref{fig:three_proteins}A. We also compare to LibINVENT \citep{Libinvent}, an established generative method for multi-vector expansion, using the implementation from the \texttt{REINVENT4} \citep{Reinvent} framework -- allocating 100K objective function calls for parity.

Fig. \ref{fig:three_proteins}A shows that SALSA achieves equal or better MPO scores to LibINVENT across all targeted tasks. SALSA consistently obtains higher-scoring molecules for both ROCS-TC and docking components compared to LibINVENT. Conversely, LibINVENT produces either similar or slightly improved QED scores. This is likely due to LibINVENT's prior, which is trained to generate fragments for spliced drug-like ChEMBL molecules, biasing towards QED which is fitted to ChEMBL data \citep{Bickerton}. However, Fig. \ref{fig:three_proteins}B demonstrates that SALSA finds far more unique high-scoring scaffolds compared to LibINVENT across all tasks. We suspect this stems from the  use of strict reaction filter penalties to promote synthesizability, sparsifying learning and making it difficult for LibINVENT to move away from its prior. This highlights an important advantage of explicitly optimizing within a targeted synthesizable chemical space. To reinforce this, in \ref{appendix:MPO_fragments} we show that SALSA still outperforms even when applied to spliced ChEMBL fragments.

\section{Conclusion}
SALSA is a sample-efficient and scalable algorithm for virtual screening in non-enumerable multi-vector spaces: our experiments highlight sample efficiency – SALSA identifies approximately 95\% of top-1K compounds after evaluating a small fraction of a 1M molecule space for both shape- and ligand-based objectives. Further, SALSA is able to consistently identify increasingly high-scoring molecules for design tasks in spaces up to 2T molecules. SALSA also enables multi-vector screening against MPOs, improving upon the output of a directly combarable generative method in LibINVENT, particularly in terms of diversity. This approach also has qualitative advantages:

\paragraph{Explicit control over chemical space}
Practitioners can easily inject expert knowledge through explicit and granular control over the target chemical space. Synthon or fragment sets can simply be filtered for desired physico-chemical properties at each vector a priori, as well as for practical considerations such as building block logistics (e.g., cost, lead-time).

\paragraph{Embedded synthetic route}
Given that each synthon is associated with a reaction, each molecule identified by SALSA comes with a predicted synthesis route which medicinal chemists can evaluate for real-world accessibility. While the enacted synthetic route may evolve in practice, providing these routes as starting points helps streamline the transition from the design to make phase, greatly aiding actionability. Further, the target space can be tuned for stricter practicality by ensuring that the associated reactions are robust and constrained to up-to-date building block catalogues.

\paragraph{Limitations and future directions} 
Successive research in this direction will likely involve iterating on the underlying modeling assumptions and acquisition strategy to more delicately balance exploration and exploitation, continuing to improve sample efficiency. It is also plausible to further increase SALSA's computational efficiency and scalability by acquiring synthons without exhaustive surrogate model inference while maintaining explicit control over the target space, e.g. by adapting the action space in existing \textit{de novo} methods such as \citet{synflownet}. Removing this constraint may also enable joint modeling of the synthon space, alleviating the implicit, naive independence assumptions that enable SALSA to scale but risk breaking down against more complex objective functions. We also expect the application of SALSA and similar algorithms to extend to molecular design tasks other than multi-vector expansion, for example to scaffold-hopping, and to screening of ultra-large synthesis-on-demand libraries, such as Enamine's REAL \citep{EnamineREAL}. We leave these considerations to future work. 

\newpage 

\bibliography{main}

\begin{thebibliography}{33}
\providecommand{\natexlab}[1]{#1}
\providecommand{\url}[1]{\texttt{#1}}
\expandafter\ifx\csname urlstyle\endcsname\relax
  \providecommand{\doi}[1]{doi: #1}\else
  \providecommand{\doi}{doi: \begingroup \urlstyle{rm}\Url}\fi

\bibitem[Bellmann et~al.(2021)Bellmann, Penner, and Rarey]{fast-similarity}
Louis Bellmann, Patrick Penner, and Matthias Rarey.
\newblock {Topological Similarity Search in Large Combinatorial Fragment Spaces}.
\newblock \emph{Journal of Chemical Information and Modeling}, 61\penalty0 (1):\penalty0 238--251, 2021.
\newblock ISSN 1549-9596.
\newblock \doi{10.1021/acs.jcim.0c00850}.

\bibitem[Bickerton et~al.(2012)Bickerton, Paolini, Besnard, Muresan, and Hopkins]{Bickerton}
G.~Richard Bickerton, Gaia~V. Paolini, Jérémy Besnard, Sorel Muresan, and Andrew~L. Hopkins.
\newblock {Quantifying the chemical beauty of drugs}.
\newblock \emph{Nature Chemistry}, 4\penalty0 (2):\penalty0 90, 2012.
\newblock ISSN 1755-4349.
\newblock \doi{10.1038/nchem.1243}.

\bibitem[Bradshaw et~al.(2019)Bradshaw, Paige, Kusner, Segler, and Hernández-Lobato]{gen-syn-mols}
John Bradshaw, Brooks Paige, Matt~J Kusner, Marwin H~S Segler, and José~Miguel Hernández-Lobato.
\newblock {A Model to Search for Synthesizable Molecules}.
\newblock \emph{arXiv}, 2019.
\newblock \doi{10.48550/arxiv.1906.05221}.

\bibitem[Carhart et~al.(1985)Carhart, Smith, and Venkataraghavan]{Carhart1985AtomPA}
Raymond~E. Carhart, Dennis~H. Smith, and R.~Venkataraghavan.
\newblock Atom pairs as molecular features in structure-activity studies: definition and applications.
\newblock \emph{J. Chem. Inf. Comput. Sci.}, 25:\penalty0 64--73, 1985.
\newblock URL \url{https://api.semanticscholar.org/CorpusID:37017771}.

\bibitem[Cheng \& Beroza(2023)Cheng and Beroza]{SASS}
Chen Cheng and Paul Beroza.
\newblock {Shape-Aware Synthon Search (SASS) for virtual screening of synthon-based chemical spaces}.
\newblock \emph{ChemRxiv}, 2023.
\newblock \doi{10.26434/chemrxiv-2023-68jzh}.

\bibitem[Cretu et~al.(2024)Cretu, Harris, Igashov, Schneuing, Segler, Correia, Roy, Bengio, and Liò]{synflownet}
Miruna Cretu, Charles Harris, Ilia Igashov, Arne Schneuing, Marwin Segler, Bruno Correia, Julien Roy, Emmanuel Bengio, and Pietro Liò.
\newblock Synflownet: Design of diverse and novel molecules with synthesis constraints, 2024.
\newblock URL \url{https://arxiv.org/abs/2405.01155}.

\bibitem[Enamine(2024)]{EnamineREAL}
Enamine.
\newblock {REAL} database, 2024.
\newblock URL \url{https://enamine.net/compound-collections/real-compounds/real-database}.

\bibitem[Falcon \& {The PyTorch Lightning team}(2019)Falcon and {The PyTorch Lightning team}]{pytorch-lightning}
William Falcon and {The PyTorch Lightning team}.
\newblock {PyTorch Lightning}, March 2019.
\newblock URL \url{https://github.com/Lightning-AI/lightning}.

\bibitem[Fan et~al.(2020)Fan, Tan, Chen, Qi, Nie, Luo, Cheng, and Wang]{D2}
Luyu Fan, Liang Tan, Zhangcheng Chen, Jianzhong Qi, Fen Nie, Zhipu Luo, Jianjun Cheng, and Sheng Wang.
\newblock {Haloperidol bound D2 dopamine receptor structure inspired the discovery of subtype selective ligands}.
\newblock \emph{Nature Communications}, 11\penalty0 (1):\penalty0 1074, 2020.
\newblock \doi{10.1038/s41467-020-14884-y}.

\bibitem[Fialkov\'a et~al.(2021)Fialkov\'a, Zhao, Papadopoulos, Engkvist, Bjerrum, Kogej, and Patronov]{Libinvent}
Vendy Fialkov\'a, Jiaxi Zhao, Kostas Papadopoulos, Ola Engkvist, Esben~Jannik Bjerrum, Thierry Kogej, and Atanas Patronov.
\newblock {LibINVENT: Reaction-based Generative Scaffold Decoration for in Silico Library Design}.
\newblock \emph{Journal of Chemical Information and Modeling}, 2021.
\newblock ISSN 1549-9596.
\newblock \doi{10.1021/acs.jcim.1c00469}.

\bibitem[Gal \& Ghahramani(2016)Gal and Ghahramani]{Gal}
Yarin Gal and Zoubin Ghahramani.
\newblock Dropout as a bayesian approximation: Representing model uncertainty in deep learning.
\newblock In \emph{Proceedings of the 33nd International Conference on Machine Learning, {ICML}}, volume~48, pp.\  1050--1059. JMLR.org, 2016.
\newblock URL \url{http://proceedings.mlr.press/v48/gal16.html}.

\bibitem[Gao \& Coley(2020)Gao and Coley]{Gao}
Wenhao Gao and Connor~W. Coley.
\newblock {The Synthesizability of Molecules Proposed by Generative Models}.
\newblock \emph{Journal of Chemical Information and Modeling}, 60\penalty0 (12):\penalty0 5714--5723, 04 2020.
\newblock ISSN 1549-9596.
\newblock \doi{10.1021/acs.jcim.0c00174}.
\newblock URL \url{https://doi.org/10.1021/acs.jcim.0c00174}.

\bibitem[Gao et~al.(2024)Gao, Luo, and Coley]{syn-gen-analogue}
Wenhao Gao, Shitong Luo, and Connor~W Coley.
\newblock {Generative Artificial Intelligence for Navigating Synthesizable Chemical Space}.
\newblock \emph{arXiv}, 2024.

\bibitem[Graff et~al.(2021)Graff, Shakhnovich, and Coley]{Graff}
David~E Graff, Eugene~I Shakhnovich, and Connor~W Coley.
\newblock {Accelerating high-throughput virtual screening through molecular pool-based active learning}.
\newblock \emph{Chem Sci.}, 2021.
\newblock ISSN 2021;12(22):7866-7881.
\newblock \doi{10.1039/d0sc06805e}.

\bibitem[Grisoni et~al.(2021)Grisoni, Huisman, Button, Moret, Atz, Merk, and Schneider]{de_novo_syn}
Francesca Grisoni, Berend J.~H. Huisman, Alexander~L. Button, Michael Moret, Kenneth Atz, Daniel Merk, and Gisbert Schneider.
\newblock {Combining generative artificial intelligence and on-chip synthesis for de novo drug design}.
\newblock \emph{Science Advances}, 7\penalty0 (24):\penalty0 eabg3338, 2021.
\newblock ISSN 2375-2548.
\newblock \doi{10.1126/sciadv.abg3338}.

\bibitem[Heid et~al.(2024)Heid, Greenman, Chung, Li, Graff, Vermeire, Wu, Green, and McGill]{chemprop}
Esther Heid, Kevin~P. Greenman, Yunsie Chung, Shih-Cheng Li, David~E. Graff, Florence~H. Vermeire, Haoyang Wu, William~H. Green, and Charles~J. McGill.
\newblock {Chemprop: A Machine Learning Package for Chemical Property Prediction}.
\newblock \emph{Journal of Chemical Information and Modeling}, 64\penalty0 (1):\penalty0 9--17, 2024.
\newblock ISSN 1549-9596.
\newblock \doi{10.1021/acs.jcim.3c01250}.

\bibitem[Klarich et~al.(2024)Klarich, Goldman, Kramer, Riley, and Walters]{Klarich}
Kathryn Klarich, Brian Goldman, Trevor Kramer, Patrick Riley, and W.~Patrick Walters.
\newblock {Thompson Sampling An Efficient Method for Searching Ultralarge Synthesis on Demand Databases}.
\newblock \emph{Journal of Chemical Information and Modeling}, 2024.
\newblock ISSN 1549-9596.
\newblock \doi{10.1021/acs.jcim.3c01790}.

\bibitem[Liphardt \& Sander(2023)Liphardt and Sander]{Liphardt}
Thomas Liphardt and Thomas Sander.
\newblock Fast substructure search in combinatorial library spaces.
\newblock \emph{Journal of Chemical Information and Modeling}, 63\penalty0 (16):\penalty0 5133--5141, 2023.
\newblock \doi{10.1021/acs.jcim.3c00290}.
\newblock URL \url{https://doi.org/10.1021/acs.jcim.3c00290}.
\newblock PMID: 37221856.

\bibitem[Loeffler et~al.(2024)Loeffler, He, Tibo, Janet, Voronov, Mervin, and Engkvist]{Reinvent}
Hannes~H. Loeffler, Jiazhen He, Alessandro Tibo, Jon~Paul Janet, Alexey Voronov, Lewis~H. Mervin, and Ola Engkvist.
\newblock {Reinvent 4: Modern AI–driven generative molecule design}.
\newblock \emph{Journal of Cheminformatics}, 16\penalty0 (1):\penalty0 20, 2024.
\newblock ISSN 1758-2946.
\newblock \doi{10.1186/s13321-024-00812-5}.

\bibitem[Luo et~al.(2024)Luo, Gao, Wu, Peng, Coley, and Ma]{analogue-search}
Shitong Luo, Wenhao Gao, Zuofan Wu, Jian Peng, Connor~W. Coley, and Jianzhu Ma.
\newblock Projecting molecules into synthesizable chemical spaces, 2024.
\newblock URL \url{https://arxiv.org/abs/2406.04628}.

\bibitem[Lyu et~al.(2023)Lyu, Irwin, and Shoichet]{Lyu_virtual_screening}
Jiankun Lyu, John~J. Irwin, and Brian~K. Shoichet.
\newblock {Modeling the expansion of virtual screening libraries}.
\newblock \emph{Nature Chemical Biology}, 19\penalty0 (6):\penalty0 712--718, 2023.
\newblock ISSN 1552-4450.
\newblock \doi{10.1038/s41589-022-01234-w}.

\bibitem[Mcule(accessed 18 Sep 2023)]{Mcule}
Mcule.
\newblock {Mcule Database}, accessed 18 Sep 2023.
\newblock URL \url{https://mcule.com/database/}.

\bibitem[Mendez et~al.(2019)Mendez, Gaulton, Bento, Chambers, De~Veij, Félix, Magariños, Mosquera, Mutowo, Nowotka, Gordillo-Marañón, Hunter, Junco, Mugumbate, Rodriguez-Lopez, Atkinson, Bosc, Radoux, Segura-Cabrera, Hersey, and Leach]{chembl}
David Mendez, Anna Gaulton, A.~P. Bento, Jon Chambers, Marleen De~Veij, Edwin Félix, María~P. Magariños, Juan~F. Mosquera, Prudence Mutowo, Michał Nowotka, Marta Gordillo-Marañón, Fiona Hunter, Leandro Junco, Grace Mugumbate, María Rodriguez-Lopez, Francis Atkinson, Nicolas Bosc, Charles~J. Radoux, Aldo Segura-Cabrera, Anne Hersey, and Andrew~R. Leach.
\newblock Chembl: towards direct deposition of bioassay data.
\newblock \emph{Nucleic Acids Res}, 47\penalty0 (D1):\penalty0 D930--D940, 2019.
\newblock \doi{10.1093/nar/gky1075}.

\bibitem[{OpenEye}(2022)]{OpenEye2022}
{OpenEye}.
\newblock {OpenEye Scientific Software}, {OpenEye} toolkits (2022.1.1).
\newblock \url{http://www.eyesopen.com}, 2022.

\bibitem[Paszke et~al.(2019)Paszke, Gross, Massa, Lerer, Bradbury, Chanan, Killeen, Lin, Gimelshein, Antiga, et~al.]{pytorch}
Adam Paszke, Sam Gross, Francisco Massa, Adam Lerer, James Bradbury, Gregory Chanan, Trevor Killeen, Zeming Lin, Natalia Gimelshein, Luca Antiga, et~al.
\newblock Pytorch: An imperative style, high-performance deep learning library.
\newblock \emph{Advances in neural information processing systems}, 32, 2019.

\bibitem[Rajapakse et~al.(2006)Rajapakse, Nantermet, Selnick, Munshi, McGaughey, Lindsley, Young, Lai, Espeseth, Shi, Colussi, Pietrak, Crouthamel, Tugusheva, Huang, Xu, Simon, Kuo, Hazuda, Graham, and Vacca]{BACE1}
Hemaka~A. Rajapakse, Philippe~G. Nantermet, Harold~G. Selnick, Sanjeev Munshi, Georgia~B. McGaughey, Stacey~R. Lindsley, Mary~Beth Young, Ming-Tain Lai, Amy~S. Espeseth, Xiao-Ping Shi, Dennis Colussi, Beth Pietrak, Ming-Chih Crouthamel, Katherine Tugusheva, Qian Huang, Min Xu, Adam~J. Simon, Lawrence Kuo, Daria~J. Hazuda, Samuel Graham, and Joseph~P. Vacca.
\newblock {Discovery of Oxadiazoyl Tertiary Carbinamine Inhibitors of $\beta$-Secretase (BACE-1)}.
\newblock \emph{Journal of Medicinal Chemistry}, 49\penalty0 (25):\penalty0 7270--7273, 2006.
\newblock ISSN 0022-2623.
\newblock \doi{10.1021/jm061046r}.

\bibitem[{RDKit}(2023)]{RDKit}
{RDKit}.
\newblock Open-source cheminformatics. (2023.09.6).
\newblock \url{https://www.rdkit.org}, 2023.

\bibitem[Renz et~al.(2019)Renz, Rompaey, Wegner, Hochreiter, and Klambauer]{Gen_failures}
Philipp Renz, Dries~Van Rompaey, Jörg~Kurt Wegner, Sepp Hochreiter, and Günter Klambauer.
\newblock {On failure modes in molecule generation and optimization}.
\newblock \emph{Drug Discovery Today: Technologies}, 32:\penalty0 55--63, 2019.
\newblock ISSN 1740-6749.
\newblock \doi{10.1016/j.ddtec.2020.09.003}.

\bibitem[Sadybekov et~al.(2022)Sadybekov, Sadybekov, Liu, Iliopoulos-Tsoutsouvas, Huang, Pickett, Houser, Patel, Tran, Tong, Zvonok, Jain, Savych, Radchenko, Nikas, Petasis, Moroz, Roth, Makriyannis, and Katritch]{Sadybekov}
Arman~A. Sadybekov, Anastasiia~V. Sadybekov, Yongfeng Liu, Christos Iliopoulos-Tsoutsouvas, Xi-Ping Huang, Julie Pickett, Blake Houser, Nilkanth Patel, Ngan~K. Tran, Fei Tong, Nikolai Zvonok, Manish~K. Jain, Olena Savych, Dmytro~S. Radchenko, Spyros~P. Nikas, Nicos~A. Petasis, Yurii~S. Moroz, Bryan~L. Roth, Alexandros Makriyannis, and Vsevolod Katritch.
\newblock {Synthon-based ligand discovery in virtual libraries of over 11 billion compounds}.
\newblock \emph{Nature}, 601\penalty0 (7893):\penalty0 452--459, 2022.
\newblock ISSN 0028-0836.
\newblock \doi{10.1038/s41586-021-04220-9}.

\bibitem[Schmidt et~al.(2021)Schmidt, Klein, and Rarey]{fast-substructure}
Robert Schmidt, Raphael Klein, and Matthias Rarey.
\newblock {Maximum Common Substructure Searching in Combinatorial Make-on-Demand Compound Spaces}.
\newblock \emph{Journal of Chemical Information and Modeling}, 2021.
\newblock ISSN 1549-9596.
\newblock \doi{10.1021/acs.jcim.1c00640}.

\bibitem[Schneider(2017)]{design_automation}
Gisbert Schneider.
\newblock {Automating drug discovery}.
\newblock \emph{Nature Reviews Drug Discovery}, 17\penalty0 (2):\penalty0 97, 02 2017.
\newblock ISSN 1474-1784.
\newblock \doi{10.1038/nrd.2017.232}.
\newblock URL \url{http://www.nature.com/articles/nrd.2017.232}.

\bibitem[Weiss et~al.(2023)Weiss, Yanes, Chakraborty, Cosmo, Bronstein, and Gershoni-Poranne]{guided_diffusion}
Tomer Weiss, Eduardo~Mayo Yanes, Sabyasachi Chakraborty, Luca Cosmo, Alex~M. Bronstein, and Renana Gershoni-Poranne.
\newblock {Guided diffusion for inverse molecular design}.
\newblock \emph{Nature Computational Science}, 3\penalty0 (10):\penalty0 873--882, 2023.
\newblock \doi{10.1038/s43588-023-00532-0}.

\bibitem[Wood et~al.(2019)Wood, Korolchuk, Tatum, Wang, Endicott, Noble, and Martin]{CDK2}
Daniel~J. Wood, Svitlana Korolchuk, Natalie~J. Tatum, Lan-Zhen Wang, Jane~A. Endicott, Martin~E.M. Noble, and Mathew~P. Martin.
\newblock {Differences in the Conformational Energy Landscape of CDK1 and CDK2 Suggest a Mechanism for Achieving Selective CDK Inhibition}.
\newblock \emph{Cell Chemical Biology}, 26\penalty0 (1):\penalty0 121--130.e5, 2019.
\newblock ISSN 2451-9456.
\newblock \doi{10.1016/j.chembiol.2018.10.015}.

\end{thebibliography}
\bibliographystyle{main}

\appendix

\newpage 

\section{Appendix}

\subsection{SALSA algorithm}\label{appendix:algorithm}

In algorithm \ref{appendix:algorithm_actual}, we present pseudo-code for SALSA as implemented in the above experiments, i.e. applied to 2-vector enumeration, with a surrogate model per vector.

\begin{algorithm}
[H]\label{appendix:algorithm_actual}
\SetArgSty{textnormal}
\SetKwInput{Where}{Where}
\SetKwInput{Parameter}{Parameters}
\SetKwInput{Config}{Config}

\SetCommentSty{textit}
\SetFuncSty{textbf}
\SetKwComment{tcp}{\# }{}
\DontPrintSemicolon
\caption{Scalable Active Learning via Synthon Acquisition}

\KwIn{Synthon sets $\mathcal{S}_i$ for $i \in \{0, 1\}$; objective function $f: \texttt{mols} \to \mathbb{R}$}

\Config{Surrogate models $\hat{f}_i(s) \to \mathcal{N}(\mu(s), \sigma(s))$ for $s \in \mathcal{S}_i$, $i \in \{0,1\}$; \newline $\texttt{N} \in \texttt{int}$ rounds; $\texttt{K} \in \texttt{int}$ samples per round; $\rho_{\texttt{max}}\in \texttt{int}$ max sample attempts; \newline acquisition strategy $\alpha_f:  \mathcal{S} \leadsto \mathbb{R}$ (i.e. stochastic e.g. for TS $\alpha_f(s) \gets x \sim f(s)$)}

\KwOut{$\mathcal{M}_{f} \subset \texttt{mols} \times \mathbb{R}$ a set of scored molecules w.r.t. $f$}\;

\tcp{Randomly sample \texttt{K} molecules}
$\mathcal{M}_{\texttt{new}} \gets \{(\texttt{mol}(s_0, s_1) \text{ for } (s_0, s_1) \in \texttt{zip}(\texttt{random}(\mathcal{S}_0), \texttt{random}(\mathcal{S}_1))\}$\; 
$\mathcal{M}_f \gets \emptyset$, $\rho \gets 0$ \tcp{Initialize scored set, and sample attempt count} \;

\For{$n \gets 1$ \KwTo \texttt{N}}{
    $\mathcal{M}_f \gets \mathcal{M}_f \cup \{(m, f(m)) \texttt{ for } m \in \mathcal{M}_{\texttt{new}}\}$  \tcp{Score new molecules}

    $\mathcal{M}_{\texttt{new}} \gets \emptyset$\;\;

    \If{$n < N$ and $\rho \leq \rho_{\texttt{max}}$}{
    
        $\mathcal{D}_i \gets \{(m_i, y) \texttt{ for } (m, y) \in \mathcal{M}_f\}$ \,\, for $i \in \{0, 1\}$ \tcp{Update synthon datasets}
        
        $f^*_i \gets \hat{f}_i\texttt{.fit}(\mathcal{D}_i)$ \,\,  for $i \in \{0, 1\}$        \tcp{Train surrogate models}\;

        \tcp{Acquire new molecules via synthons}
        \tcp{$\rho$ will terminate loop if new molecules are sampled too infrequently (i.e. convergence)}
        $\rho \gets 0$\;
        
        \While{$\texttt{len}(\mathcal{M}_\mathcal{\texttt{new}}) < \texttt{K} \texttt{ and count} < \rho $}{
            $s^*_i \gets \operatorname{argmax}\limits_{s \in \mathcal{S}_i}\alpha_{f^*_i}(s) \text{ for } i \in \{0, 1\}$\;
            $\rho \gets \rho + 1$\;  
            \If{$\texttt{ mol}(s_0, s_1) \not\in \mathcal{M}_{\texttt{new}} \cup \{m \text{ for } (m, \_) \in \mathcal{M}_{f}\}$}{
                        $\mathcal{M}_\texttt{new}\texttt{.add}(\texttt{mol}(s_0, s_1))$   \tcp{Only add sampled molecule if unseen}
            }
        }
    }   
}

\Return $\mathcal{M}_f$\;\;
\Where{
    $\texttt{mol}: \mathcal{S}_0 \times \mathcal{S}_1 \to \texttt{mols} \text{ and for convenience } \texttt{mol}(s_0, s_1)_i := s_i$\
}
\end{algorithm}

\subsection{Synthon spaces}\label{appendix:synthon_spaces}

\begin{figure}[H]
    \centering\includegraphics[scale=0.3]{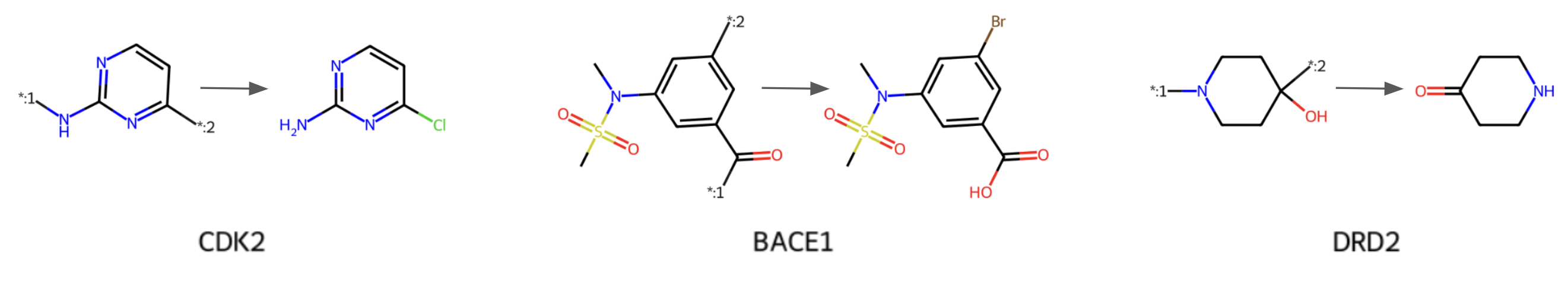}
    \captionsetup{labelfont=bf}
    \caption{Core scaffolds mapped to synthetic intermediates with functionalized reaction handles.}
    \label{fig::scaffolds}
\end{figure}

\paragraph{Molecular scaffolds}
In our experiments, SALSA was applied in the context of multi-vector enumeration from an explicit core. We extracted cores from ligands found in pertinent PDB structures containing at least one ring system with two R-groups for optimization, specifically from 3D co-crystallised systems in order to facilitate shape- and structure-based scoring. We identified CDK2 (PDB: 6GUH)\citep{CDK2}, BACE1 (PDB: 2IRZ) \citep{BACE1}, and DRD2 (PDB: 6LUQ) \citep{D2} as suitable candidates to represent design tasks. Each extracted core was replaced with a synthetic intermediate with functionalized reaction handles at the target R-groups to enable synthon generation for our experimental target spaces (see Fig. \ref{fig::scaffolds}).

\paragraph{Synthon space construction}
For our experiments, we explicitly construct large multi-vector spaces. Given a core scaffold with R-group handles at desired vectors, we determine applicable reactions via partial substructure matching over a set of bimolecular SMIRKS reactions. Building blocks that match the corresponding pattern are retrieved from a database of commercially available options via \citet{Mcule}. Synthons are created by replacing displaced reacting groups with a generic linker atom, as detailed in \citet{Liphardt}.

\subsection{MPNN hyperparameters}\label{appendix:MPNN_hyperparameters}

We minimally adapt the default \texttt{chemprop} \citep{chemprop} architecture: a message-passing depth of \texttt{3} for the encoder, \texttt{2} layers for the MVE head, and \texttt{300} hidden dimensions with ReLU activations throughout for both. In each round, we trained from scratch for a maximum of \texttt{50} epochs with a batch size of \texttt{64}, holding out \texttt{20\%} data for early-stopping on the MVE validation loss with patience\texttt{=10}. We optimized with Adam and a NoamLR scheduler with initial, max, and final LRs of \texttt{1e-4}, \texttt{1e-3}, \texttt{1e-4}, respectively. The MPNN itself is implemented in \texttt{PyTorch} \citep{pytorch} and trained via \texttt{lightning} \citep{pytorch-lightning}.

\subsection{Alternative acquisition strategies and surrogate models}\label{appendix:alternative_acquisition_strategies}
\begin{figure}[H]
    \centering
    \includegraphics{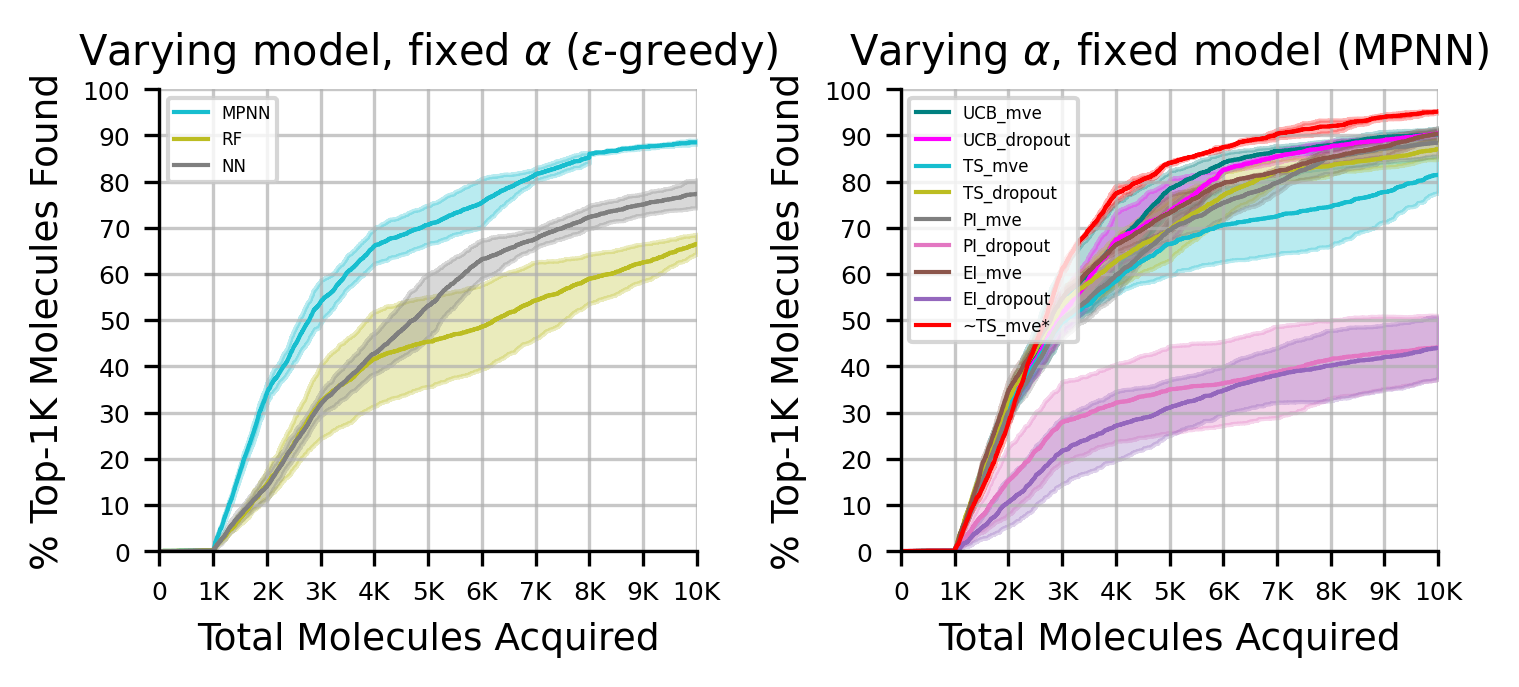}
    \captionsetup{labelfont=bf}
    \caption{\textbf{Ablation}   
    Acquisition of the top-1K scoring compounds on the 1M-space ROCS-TC design task. Each method was allocated a 1K objective budget per round for 10 rounds.}  \label{fig::acquisition_ablation}
\end{figure}

In Fig. \ref{fig::acquisition_ablation} we ablate MolPAL's implementation of random forest (RF) and feed-forward neural network (NN) models against the MPNN model used in our experiments, using 2048-bit atom-pair fingerprints with a minimal and maximal radius of 1 and 3 \citep{Carhart1985AtomPA} as fixed features. Here, the acquisition strategy is held fixed to $\epsilon$-greedy, where top synthons are chosen except for an $\epsilon=5$\% chance of picking at random. We see that the MPNN significantly outperforms the NN and RF. We also report performance of a number of non-stochastic acquisition strategies used in MolPAL, including probability of improvement (PI), expected improvement (EI), upper confidence bounds (UCB), and a ``non-stochastic'' TS where scores are drawn only once (see \citet{Graff} for definitions). Throughout, a molecule's acquisition score was defined as the sum of its synthon acquisition scores. We found that stochastic TS performed best. This is consistent with the promising performance of the method in \citet{Klarich}, which is conceptually similar to this configuration of SALSA, instead using tabulated predictions with online TS. This suggests that optimistic, exploratory strategies are most effective when navigating via synthon spaces in this manner. 

We also trialed inference-time dropout as an alternative to MVE for uncertainty quantification, training to predict the mean via MSE loss, and estimating the mean and variance by sampling \texttt{10} predictions with a dropout probability of \texttt{0.2}. This method can be interpreted as an approximate Bayesian (i.e. variational) technique for modeling epistemic uncertainty \citep{Gal}. We believe the drastically inferior performance of this approach when compared to its application in \citet{Graff} stems from the aleatoric variance of a given synthon's score distribution dominating the epistemic variance in its predicted mean, due to the unobserved choice of complementary synthon.

\subsection{SALSA with One Model}\label{appendix:one_model_vs_two}

\begin{figure}[h]
    \centering
    \includegraphics[width=1\textwidth]{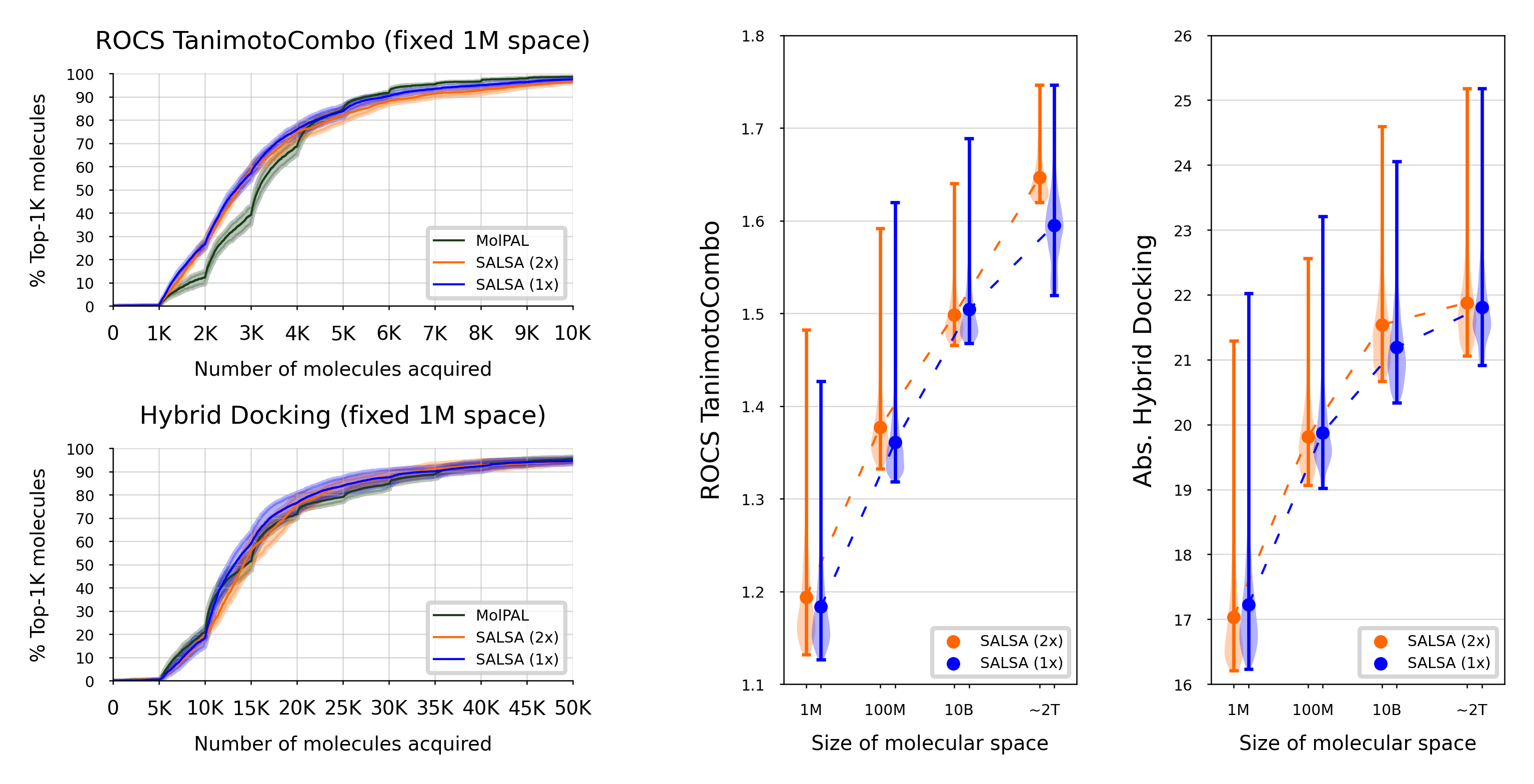}
    \captionsetup{labelfont=bf}
    \caption{One vs two-model SALSA for sample efficiency (5 trials) and scaling (3 trials)}
    \label{fig:one_v_two}
\end{figure}

In Fig. \ref{fig:one_v_two}, we repeat our sample efficiency and scaling experiments from \ref{section:results} using a single model configuration. Here, we simply add a one-hot encoding to the learned graph embedding to indicate the vector to which each synthon belongs, and the model sees batches of data from both vectors during training. We did not modify the model architecture or training hyperparameters. We find that the performance of the single-model variant is similar or slightly improved in the benchmark space, ultimately recalling a mean of 97.5\% and 94.6\% of the top-1K molecules, compared to 96.5\% and 94.5\% for two models with respect to ROCS-TC and Hybrid Docking (MolPAL: 98.5\% and 95.4\%). The scaling experiments show comparable trends for both variants as the size of the molecular space increases, but SALSA (1x) appears to perform slightly worse in the largest space for ROCS-TC. We hypothesize that the single model variant may saturate faster, requiring greater capacity to handle its larger data distribution. We highlight the possibility of using a single model due to its advantages: simpler, more efficient use of computational resources, and transferablility to other molecular design contexts such as de novo design, which may additionally benefit from more informative featurization and generalization across different synthon sets.

\subsection{Baselines}\label{appendix:baselines}

\paragraph{MolPAL}

In our sample efficiency experiments, we run MolPAL in its best reported configuration in \citet{Graff}, with an MPNN surrogate model configured as in \ref{appendix:MPNN_hyperparameters}, and greedy acquisition.

\paragraph{(Tabular) Thompson sampling}\label{methods:Thompson} 
We modified the open source implementation of \citet{Klarich} at \url{https://github.com/PatWalters/TS}, adding a custom CSV evaluator class to enable scoring with pre-computed scores. We provided the same synthon sets used by SALSA as the reagent lists. We allowed for 3 and 2 warm-up trials for Hybrid Docking and ROCS-TC scoring, respectively. We did not count the warm-up trials towards the objective function budget.

\paragraph{LibINVENT}

In our MPO experiments, we configured the LibINVENT implementation in \citet{Reinvent} in staged learning mode, running for \texttt{1600} iterations (\texttt{max\_steps}) with a \texttt{batch\_size} of \texttt{64} to generate 100K molecules. We used default $\sigma$\texttt{=128} for the Difference between Augmented and Posterior (DAP) reward strategy, and a learning rate of \texttt{1e-4}.

\subsection{Molecular Scoring Functions}\label{appendix:scores}

\paragraph{Docking: OpenEye Hybrid Docking} 

Prior to docking, sampled molecules undergo preparation, including protomer, stereochemistry, and conformer generation using \texttt{openeye-toolkits} (version 2022.1.1) \citep{OpenEye2022}. Protomer generation was performed using OpenEye Quacpac. Stereochemistry enumeration and conformer generation were carried out using OpenEye Omega, with a maximum of 10 stereocenters and 200 conformers per molecule. Following molecular preparation, docking was conducted using OpenEye Hybrid. For CDK2, BACE1, and DRD2, the outputted scores were divided by a factor of -24, -17, and -24, to scale roughly between 0 and 1. The scaled docking score of the highest scoring conformer was assigned to the molecule.

\paragraph{ROCS-TC: OpenEye TanimotoCombo Score}
For ROCS-TC, all molecules undergo preparation similarly to docking. We use OpenEye ROCS to perform shape similarity scoring. We again define a molecule's score to be the highest TanimotoCombo Score achieved by any of its conformers, where TanimotoCombo Score consists of an equally weighted sum of shape and colour Tanimoto scores. We divided these scores by 2 during learning, to scale the output range between 0 and 1.

\paragraph{QED}
QED \citep{Bickerton} was calculated using RDKit \citep{RDKit}. Each enumerated molecule was converted into an RDKit molecule and scored using the RDKit QED function. 

\paragraph{MPO objective functions}

We defined two simple linear MPOs for each of the three protein targets – CDK2, BACE1, and DRD2. ROCS-TC + QED was weighted 2:1. In Docking + QED, QED was weighted equally to the scaled docking scores.

\subsection{ChEMBL molecules, Physio-chemical calculations and ADMET predictions}\label{appendix:ADMET}
\begin{figure}[H]
    \centering
    \includegraphics[width=0.95\textwidth]{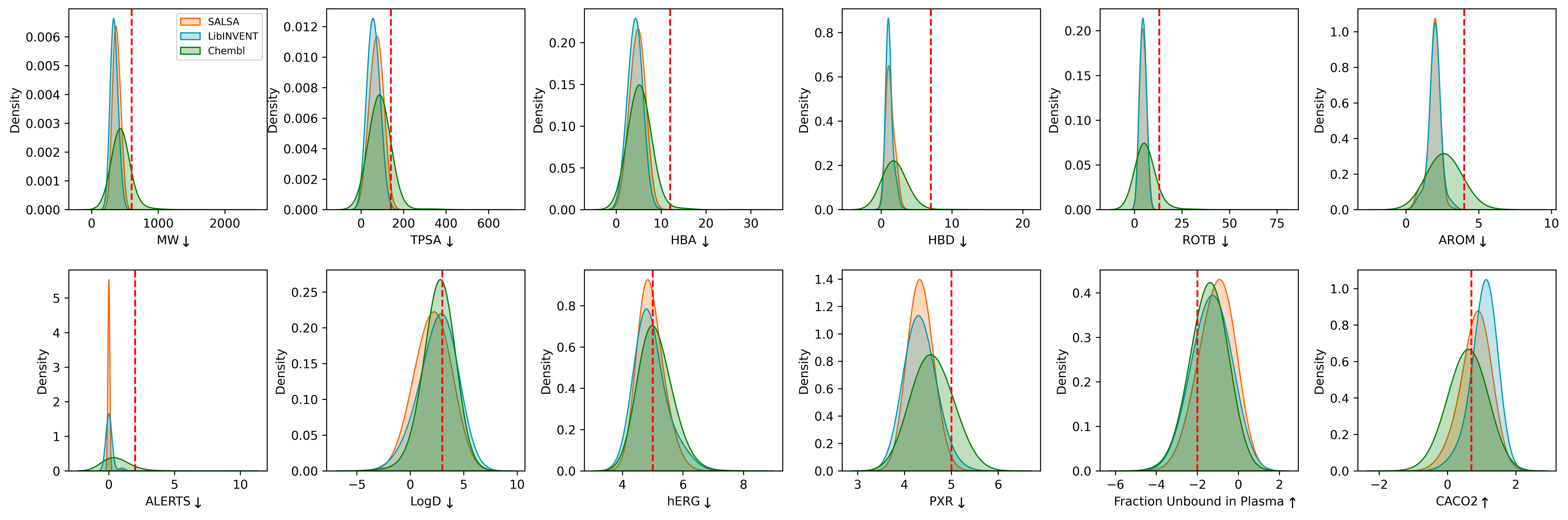}
    \captionsetup{labelfont=bf}
    \caption{
    \label{fig:ADMET}ADMET properties for the top-1K molecules generated by SALSA and LibINVENT compared with the top-1K bioactive molecules from ChEMBL for each protein target, ranked by pChEMBL value. Dashed lines represent thresholds for these properties, and an up or down arrow represents preference for values greater or less than the threshold, respectively. Density estimates are aggregated over all MPOs and protein targets to give a high-level view on property distributions.
    }
    \label{fig:admet}
\end{figure}

In Fig. \ref{fig:ADMET}, the top-1K  molecules from SALSA and LibINVENT runs are compared to the top-1K bioactive molecules in ChEMBL as measured by pChEMBL score for each protein target. Properties were calculated via \citet{RDKit} where possible, or otherwise predicted (via internal models) in order to assess drug-likeness over twelve metrics: molecular weight (MW), total polar surface area (TPSA), hydrogen bond acceptors (HBA), hydrogen bond donors (HBD), number of aromatic rings (AROM), number of structural alerts (ALERTS), predicted lipophilicity at pH 7.4 (LogD), number of rotatable bonds (ROTB), hERG potency (hERG), PXR potency (PXR), log fraction of ligand unbound in human plasma (Fraction Unbound in Plasma), and CACO2 permeability (CACO2). We observe the majority of molecules generated by both methods fall within desirable drug-like space and within similar or better bounds than molecules retrieved from ChEMBL.

\paragraph{ChEMBL} Bioactive molecules associated with each protein target were downloaded from the ChEMBL33 database \citep{chembl}. Molecules without associated SMILES values were removed. The top-1K (unique) molecules with highest pChEMBL values for each target were selected, where pChEMBL value is the negative logarithm of the molar IC50, EC50, Ki, Kd, or Potency.

\subsection{Runtime and model inference}\label{appendix:runtime}

SALSA is able to effectively screen large multi-vector spaces within a matter of hours using a single A10G GPU for training and inference. Time spent on training, scoring, and acquisition is primarily a function of the number of total molecules acquired and the number of rounds. Model inference scales linearly with the size of the targeted synthon/fragment sets. Time spent on scoring, model training, inference, and acquisition for different space sizes in our scaling experiments in \ref{experiments:sample_efficiency} is outlined in Tables \ref{tab:rocs_time_table} and \ref{tab:docking_time_table}. Here, inference time remains negligible until the $\sim$2T space, where it begins to take $\sim$2 hours to compute the necessary statistics for each synthon. This is significant, but it is a million-fold improvement over an equivalent full-molecular pool-based screen requiring one forward pass per molecule. Inference times can easily be improved by distributed compute.

\begin{table}[ht]  
\caption{ROCS-TC Results}  
\label{tab:rocs_time_table}  
\renewcommand{\arraystretch}{2}  
\centering  
\resizebox{\textwidth}{!}{%
\begin{tabular}{lcccc}  
\hline  
                & 1M space               & 100M space              & 10B space            & 2T space               \\
\hline
scoring         & 2H 28m 41s $\pm$ 6m 25s   & 1H 41m 43s $\pm$ 6m 24s   & 1H 40m 15s $\pm$ 18m 36s  & 1H 23m 59s $\pm$ 10m 31s  \\
training        & 2H 21m 37s $\pm$ 22m 22s  & 2H 13m 42s $\pm$ 3m 48s   & 2H 15m 38s $\pm$ 4m 9s    & 2H 6m 15s $\pm$ 1m 20s    \\
inference       & 0H 0m 11s $\pm$ 0m 0s     & 0H 0m 59s $\pm$ 0m 1s     & 0H 8m 36s $\pm$ 0m 15s    & 2H 15m 46s $\pm$ 3m 52s   \\
overall         & 5H 11m 43s $\pm$ 20m 16s  & 4H 5m 0s $\pm$ 4m 36s     & 4H 20m 55s $\pm$ 20m 30s  & 8H 28m 7s $\pm$ 9m 31s    \\
\hline  
\end{tabular}%
}  
\end{table}  

\begin{table}[H]
\caption {Hybrid Docking Results} \label{tab:docking_time_table} 
\renewcommand{\arraystretch}{2}  
\centering  
\resizebox{\textwidth}{!}{%
\begin{tabular}{lcccc}  
\hline  
                & 1M space               & 100M space              & 10B space             & 2T space               \\
\hline
scoring         & 4H 3m 14s $\pm$ 18m 15s   & 3H 25m 49s $\pm$ 12m 32s  & 3H 41m 19s $\pm$ 5m 10s   & 3H 24m 14s $\pm$ 11m 59s  \\
training        & 2H 46m 46s $\pm$ 4m 43s   & 2H 36m 32s $\pm$ 9m 10s   & 2H 41m 24s $\pm$ 9m 1s    & 2H 37m 13s $\pm$ 6m 42s   \\
inference       & 0H 0m 13s $\pm$ 0m 1s     & 0H 0m 57s $\pm$ 0m 1s     & 0H 8m 45s $\pm$ 0m 11s    & 2H 21m 0s $\pm$ 4m 12s    \\
overall         & 7H 0m 9s $\pm$ 17m 9s     & 6H 12m 12s $\pm$ 8m 55s   & 6H 47m 52s $\pm$ 5m 2s    & 10H 45m 16s $\pm$ 4m 58s  \\
\hline  
\end{tabular}%
}  
\end{table}

\subsection{Top scoring molecules visualized}
\begin{figure}[H]
    \centering
    \includegraphics[width=1\textwidth]{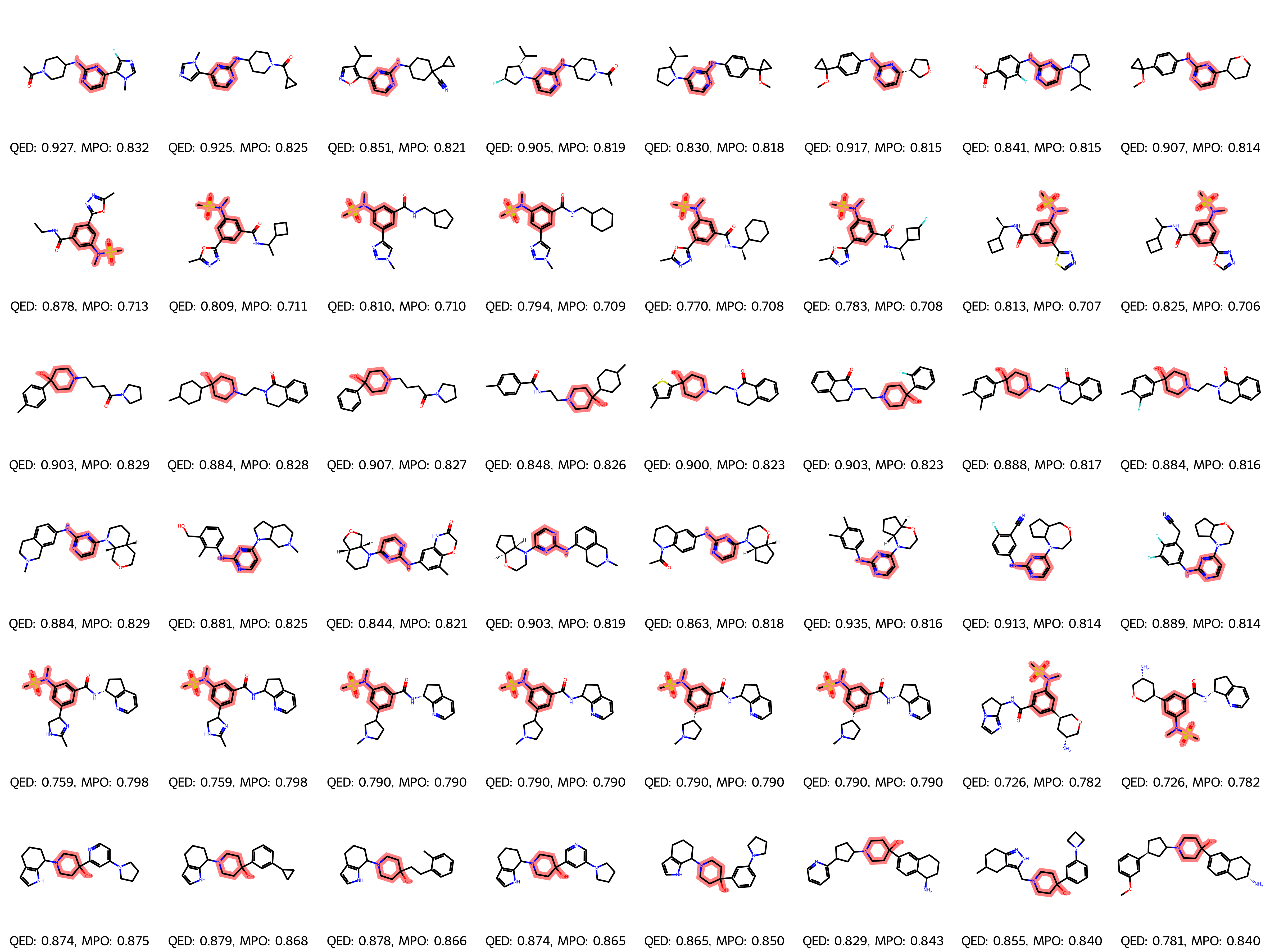}
    \captionsetup{labelfont=bf}
    \caption{Selected top molecules from SALSA MPO runs for all three protein targets. Highlighted atoms and bonds represent fixed cores. QED and MPO score are labeled below each molecule.
    }
\end{figure}

\subsection{MPOs in ChEMBL fragment space}\label{appendix:MPO_fragments}

\begin{figure}[h]
    \centering
    \includegraphics[width=1\textwidth]{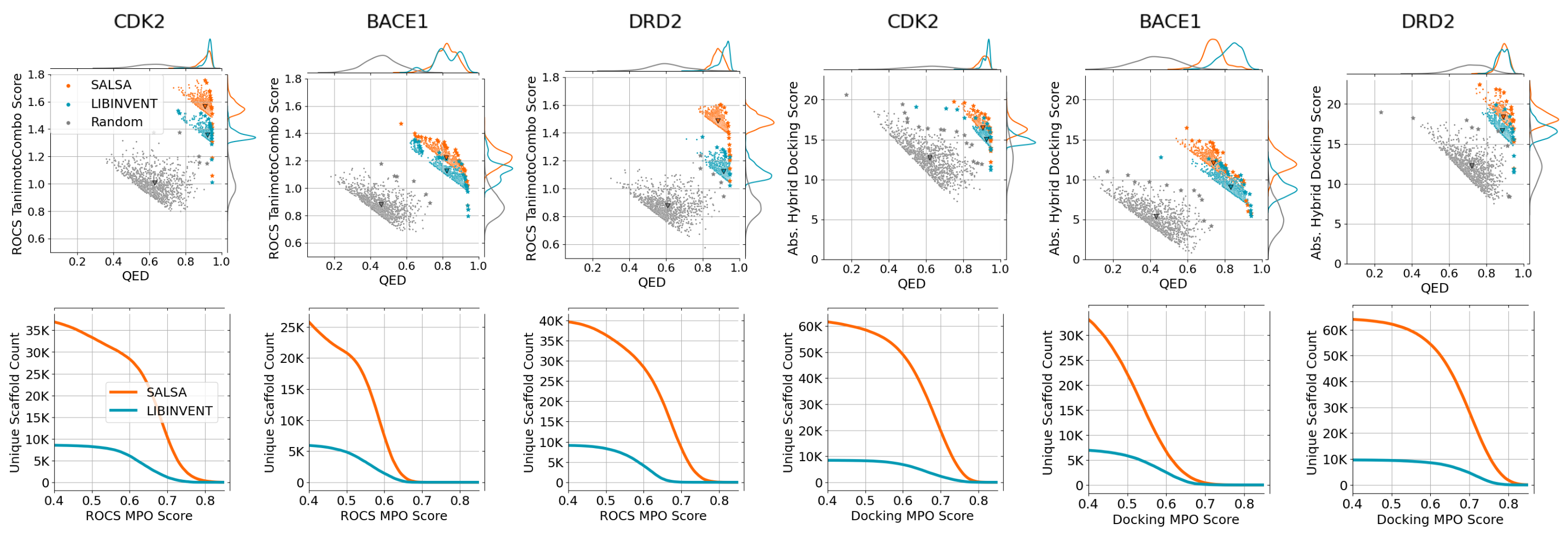}
    \captionsetup{labelfont=bf}
    \label{fig:mpo_fragments}
    \caption{Top: the top-1K molecules identified across three targets for SALSA and LibINVENT using QED \citep{Bickerton} plus ROCS-TC and Hybrid Docking objectives. Bottom: the number of unique scaffolds scoring above a given score for molecules enumerated by SALSA and LibINVENT using ROCS/QED MPO (left) and Docking/QED MPO (right) for each protein target. These runs were performed on fragment spaces formulated from ChEMBL which were generated by breaking all acyclic bonds, and then filtering for $\leq 20$ heavy atoms, $\leq 4$ H-bond donors, $\leq 4$ H-bond acceptors, $\leq 2$ rotatable bonds, $\leq 1$ ring system, and an observed frequency count of $\geq 10$.}
\end{figure}

\end{document}